%% file: root.tex
\documentclass[letterpaper, 10 pt, conference]{ieeeconf}

\IEEEoverridecommandlockouts                              

\usepackage{lineno}
\usepackage{subfigure}
\usepackage{url}
\usepackage{amsmath}
\usepackage{mathptmx}
\usepackage{amssymb}
\usepackage{mathtools}
\usepackage{multirow}
\usepackage{multicol}
\usepackage{changepage}
\usepackage{algorithm}
\usepackage{algorithmic}
\usepackage{balance}
\usepackage{anyfontsize}
\usepackage{graphicx}
\usepackage{caption}
\modulolinenumbers[5]
\usepackage{xcolor}
\usepackage{amsmath}
\usepackage{bigints}

\usepackage{array}
\newcolumntype{P}[1]{>{\centering\arraybackslash}p{#1}}

\DeclareMathAlphabet{\mathbbmsl}{U}{bbm}{m}{sl}

\usepackage{multirow}

\AtBeginDocument{%
  \providecommand\BibTeX{{%
    \normalfont B\kern-0.5em{\scshape i\kern-0.25em b}\kern-0.8em\TeX}}}

\title{\LARGE \bf
Registration between Point Cloud Streams and Sequential Bounding Boxes via Gradient Descent
}

\author{Xuesong Li$^{1}$, Xinge Zhu$^{2}$, Yuexin Ma$^{3}$, Subhan Khan$^{4}$, Jose Guivant$^{5}$ 
\thanks{$^{1}$ \small Xuesong Li is with Australia Nation University, Canberra ACT {\tt\small xuesong.li@anu.edu.au} (corresponding author)}%
\thanks{$^{2}$ \small Xinge Zhu is with the Chinese University of Hong Kong.} 
\thanks{$^{3}$ \small Yuexin Ma is with ShanghaiTech University, Shanghai {\tt\small mayuexin@shanghaitech.edu.cn}}%
\thanks{$^{4}$ \small Subhan Khan is with University of Sydney, Sydney.} 
\thanks{$^{5}$ \small Jose Guivant is with the University of New South Wales, Sydney. {\tt \small jose.guivant@unsw.edu.au}}
}

\begin{document}

\maketitle
\thispagestyle{empty}
\pagestyle{empty}

\begin{abstract}



In this paper, we propose an algorithm for registering sequential bounding boxes with point cloud streams. Unlike popular point cloud registration techniques, the alignment of the point cloud and the bounding box can rely on the properties of the bounding box, such as size, shape, and temporal information, which provides substantial support and performance gains. Motivated by this, we propose a new approach to tackle this problem. Specifically, we model the registration process through an overall objective function that includes the final goal and all constraints. We then optimize the function using gradient descent. Our experiments show that the proposed method performs remarkably well with a 40\% improvement in IoU and demonstrates more robust registration between point cloud streams and sequential bounding boxes

\end{abstract}


\section{\uppercase{Introduction}}\label{sec:introduction}
\input{introduction}

\section{\uppercase{Related work}}\label{sec:Related_work}
\input{literature_review}

\section{Problem definition}\label{sec:problem} 
\input{problem_definition}

\section{Registration}\label{sec:registration}
\input{registration}

\section{Experimental results}\label{sec:experiments}
\input{experiments}

\section{Conclusion}\label{sec:conclusion}

Overall, the paper proposes a new approach to solve the problem of registration between sequential bounding boxes and point cloud streams. The approach treats registration as an optimization problem and builds an objective function that encodes the registration goal and constraints. The experimental results with simulated data show that the approach can achieve accurate registration with a 40\% gain in IoU for 2D bounding boxes and 3D bounding boxes. However, the approach's performance decreases as the dimension of parameters increases, and testing on real-world datasets is left for future work. The paper focuses on explaining the approach's idea and demonstrating its effectiveness with simulated data.

\addtolength{\textheight}{-12cm}


\bibliographystyle{IEEEtran}
\bibliography{IEEEabrv,conference}

\end{document}

%% file: introduction.tex

\begin{figure}[!t]
    \centering
    \includegraphics[width=0.4\textwidth]{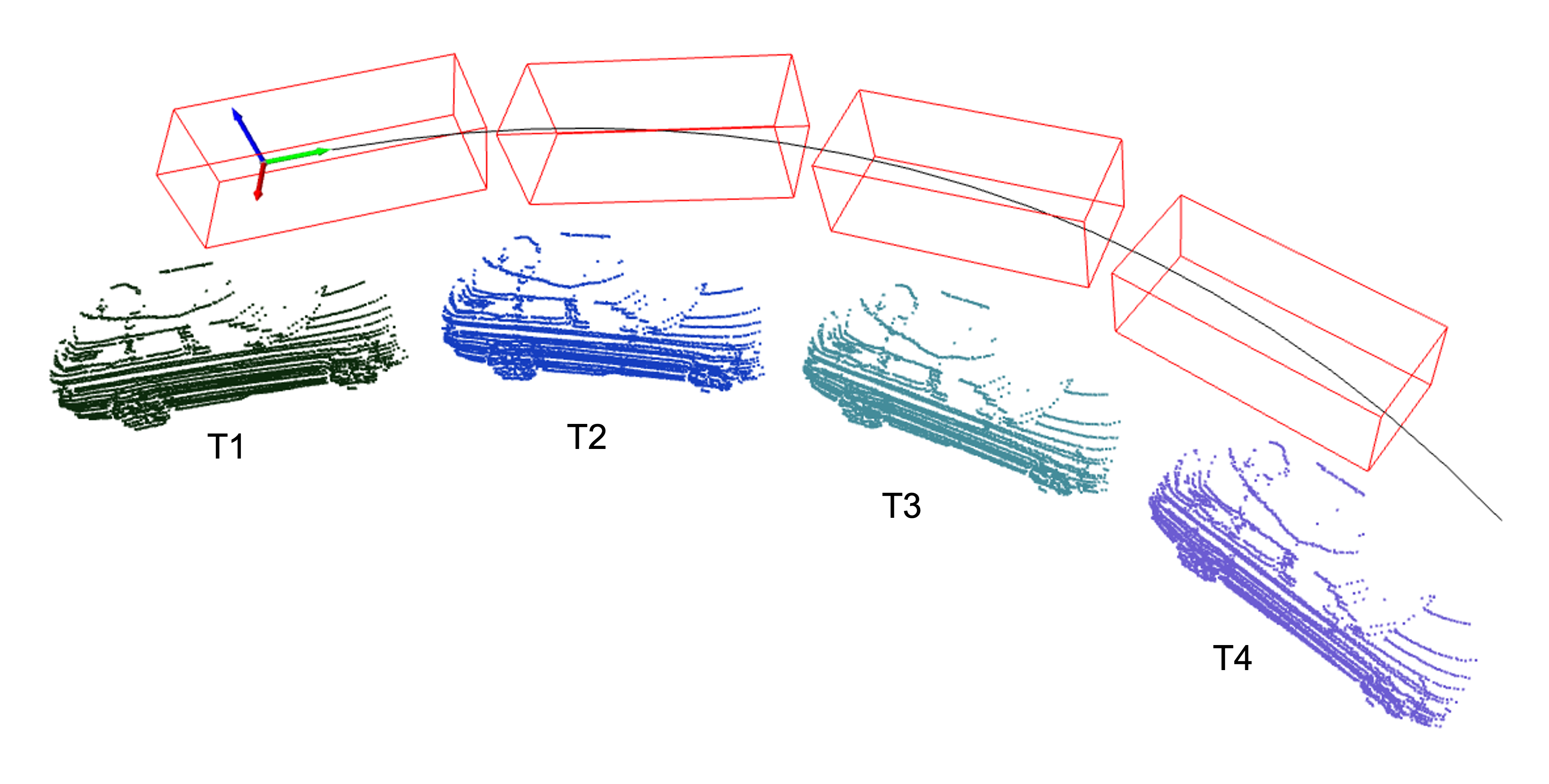}
    \caption{Point cloud streams (point clouds of ``car'' at different times) and their corresponding sequential bounding boxes (red boxes). The registration task is to align sequential bounding boxes to the given point cloud streams.}
    \label{fig:demo_prob}
\end{figure}

Point clouds are three-dimensional datasets that accurately represent the geometric location and structure of objects or surroundings. Due to their precise nature, they have become widely used across various industries, including mining, agriculture, and autonomous driving \cite{Geiger2012CVPR, nuscenes, auto4d_for_autolabelling, li2019detection, li2023efficient}. The registration between point cloud sets, which describe the same object from different points of view, has been an active research topic for a while and is typically solved using iterative closest points (ICP) and its variants. However, beyond the surface details captured by the point cloud, bounding boxes focus more on representing object instances, as they are low-dimensional and can accurately represent objects' key high-level information. Given the initial sequential bounding boxes, accurately registering and aligning them with point cloud streams, as shown in Fig. \ref{fig:demo_prob}, is still an open and new question to ask. Accurate registration between point clouds and bounding boxes is important for many real-world applications, such as path-planning tasks~\cite{liu2017path, schwarting2018planning, khan2022design} in the context of autonomous driving, object detection~\cite{shi2020pv, chen2022focal, li2020real, li2020object}, and 3D auto-labeling systems \cite{auto4d_for_autolabelling, qi2021offboard}.


One current strategy to tackle this problem is to design a multi-stage network to refine the initial bounding boxes with point clouds \cite{auto4d_for_autolabelling, qi2021offboard}. However, this approach is limited by the availability of large, high-quality annotated training datasets, which are expensive to obtain. Moreover, its performance significantly decreases when there is a domain gap between the training and deployment scenarios, and accurate alignment or registration is challenging to explain and control. Another approach is to apply the ICP algorithm by sampling points from the bounding boxes and converting this problem into the registration between point clouds. However, points sampled from bounding boxes do not have the structure of the target objects and are difficult to register accurately with the object's point set. In this paper, we propose a novel approach that borrows ideas from classical ICP algorithms and generates the solution through modeling and optimization instead of labeling and training. Similar to the ICP algorithm, we first define the final objective function and its constraints, design and verify each term in the objective function and constraints, and then combine all terms into a differentiable objective function with Lagrange multipliers. Finally, we optimize the objective function using Newton's method (second-order gradient descent). In the experimental section, we show that our approach is effective and accurate in solving the registration between point cloud streams and sequential bounding boxes.

As for as we know, this is the first approach to solve this problem via conventional methods without requiring annotated dataset. Compared with learning-based methods, the proposed method is more flexible, explained, and controllable, since we can easily change the final objective function or add extra constraints to make it adapt to different scenarios. To sum up, our main contribution is to propose a new method for solving the registration between point cloud streams and sequential bounding boxes and verifying its effectiveness and accuracy with empirical experiments.  

The rest of the paper is organized as follows. Section \ref{sec:Related_work} introduces the related work, followed by Section \ref{sec:problem} which defines the problem. Modeling and optimization can be found in Section \ref{sec:registration}. Experiments of the proposed method are described in Section \ref{sec:experiments}. Section \ref{sec:conclusion} concludes our work.

%% file: literature_review.tex

There are two main categories of algorithms for handling registration: those that use learning techniques \cite{auto4d_for_autolabelling, qi2021offboard} to train registration models, and those that use an optimization-based approach \cite{icp_original, efficient_icp, colored_icp, 6631104, joint-temporal-ops, jaremo18a} to build an objective function and optimize the distance between two registered targets. In this section, we will briefly review both approaches.

\subsection{Registration with learning}

Registration methods that use learning techniques usually employ specialized networks to gradually register bounding boxes with point clouds. For instance, in \cite{auto4d_for_autolabelling}, two separate branches were proposed to refine different attributes of 3D bounding boxes: size and dynamic pose. The first branch is designed for the size of each object, as the size will be used to crop the point clouds and is important for subsequent tasks. The second branch aims to recover and refine the pose of the object for each time point in the trajectory, given the temporal states of the entire trajectory and the point cloud (cropped with the refined size). In contrast, \cite{qi2021offboard} proposed two separate networks, based on PointNet \cite{Poinetnet}, for the static and dynamic objects, respectively. The network takes sequential point clouds and sequential bounding boxes as inputs and directly refines all attributes of 3D bounding boxes. Both methods serve as auto-labeling systems for 3D autonomous driving. These modules require higher quality training annotations than the end-to-end 3D detectors \cite{MVF_plus_plus, 3D_detector_sparecnn}, as they target an intersection over union (IoU) of 0.9 with the ground truth rather than the 0.7 IoU targeted by other detectors. Furthermore, the annotation rules in the 3D auto-labeling system can only be implicitly encoded into the labeled training samples and cannot be explicitly added into these specialized models. In comparison, our proposed method has the advantage of not requiring labeled training samples, easily embedding annotation rules, and being explainable with respect to the refined 3D bounding boxes.

\subsection{Optimization-based registration}

The most popular optimization-based registration methods are the ICP \cite{icp_original} and their variants \cite{efficient_icp, ICP_sgd, colored_icp}, which minimize the distance between corresponding points in two point clouds to register them. The algorithm iteratively selects matched points, estimates the transformation parameters by minimizing the distance between corresponding points, and transforms the one of point sets. This process is repeated until the transformation parameters converge to a stable value. However, conventional ICP is computationally expensive, and several ICP variants \cite{efficient_icp, ICP_sgd} have been proposed to speed up the algorithm. For example, \cite{efficient_icp} optimized the speed by using uniform sampling of the space of normals, and \cite{ICP_sgd} applied stochastic gradient descent to remove inefficient iterative steps. Additionally, Colored ICP \cite{colored_icp} design a photometric objective function to aligns RGB-D images to point clouds by locally parameterizing the point cloud with a virtual camera. The ICP cannot be directly applied for the registration between point cloud streams and sequential bounding boxes. The bounding box has first been converted into point set via surface sampling, then the sampled points can be registered with object's point cloud. However, sampled points lack of object's structure and geometrical information, which leads to poor performance. Our proposed method is inspired by ICP but optimizes the parameters of sequential bounding boxes directly without the need for point sampling.

%% file: problem_definition.tex
We first review the ICP algorithm \cite{icp_original}, which is a widely-used approach for the registration of point clouds. Assume that there are two corresponding sets, i.e. source points $\textit{X}$ and target points $\textit{Y}$. $\textit{X}=\{{p}_1^x, {p}_2^x, ..., {x}_n^x\}$, $\textit{Y}=\{{p}_1^y, {p}_2^y, ..., {p}_n^y\}$, in which ${p}_i^x, {p}_{i}^{y} \in \mathbbmsl{R}^{3} $. The objective of the ICP algorithm is to estimate the optimal translation $T \in \mathbbmsl{R}^{3}$ and rotation matrix $R \in \mathbbmsl{R}^{3\times3}$ that minimize the distance error between two point sets, as given in Equ.\ref{equ:icp}, where $N$ is the index set of matched point pairs. The first step is to find the matched pairs or correspondence between two point clouds, and then Equ.\ref{equ:icp} can be solved with the least square to find the $\textit{R}$ and $\textit{T}$ given the matched points. The ICP algorithm iteratively implements these two steps to estimate the optimal $\textit{R}$ and $\textit{T}$. 

\begin{equation}
\begin{aligned}
\underset{R,T}{\arg\min} \frac{1}{N} \sum_{i \in N}|| \textit{R}*{p}_i^x - \textit{T} - {p}_i^y||_{2}
\label{equ:icp}
\end{aligned}
\end{equation}


In contrast to ICP, our approach aims to align point cloud streams with their sequential bounding boxes. The source sequential bounding boxes represent the bounding boxes of a rigid object moving over a certain period. These bounding boxes have the same size (i.e. length, width, and height) but different poses (i.e. $x, y, z, roll, pitch,$ and $yaw$), and the trajectory of the central point in sequential bounding boxes should be smooth without any zigzag, as shown in Fig. \ref{fig:demo_prob}. The target point cloud streams represent the corresponding points of this rigid object at different temporal timestamps. The number of points can vary, and these points may reveal different parts of the object due to movement and occlusion.

Let's assume the sequential bounding boxes are $\textit{BOX} = \{ \textit{box}_i = [x_{i}, y_{i}, z_{i}, l, w, h, \alpha_{i}, \beta_{i}, \gamma_{i}]^T | i \in N \} $, where $(x_{i}, y_{i}, z_{i})$ denotes the central point location of the $\textit{box}_i$, $(l, w, h)$ is its size, and $(\alpha_{i}, \beta_{i}, \gamma_{i})$ indicates the orientation (roll, pitch, yaw). Given these parameters, the six planes and eight corner points of $\textit{box}_i$ can be easily calculated. The point cloud streams can be described as $\textit{SEQ}=\{ \textit{P}_{i} =\{ p^i_1, p^i_2, ..., p^i_m \}| i\in N,  p^i_j \in {\mathbbmsl{R}}^{3} \}$, where each $p^i_j$ represents the j-th 3-dim point in i-th point cloud. Our aim for the registration between point cloud streams and sequential bounding boxes is to make initial sequential bounding boxes well-aligned with the point cloud streams so that each bounding box can enclose its point clouds tightly, and simultaneously, the trajectory of all bounding boxes should be smooth. The objective of our problem is to optimize the ${\textit{R}}_i$ and ${\textit{T}}_i$ for each bounding box to minimize the distance between the points and bounding box, as Equ. \ref{equ:main_objective_old}. 

\begin{equation}
\begin{aligned}
\underset{R_{i},t_{i}}{\arg\min} \frac{1}{N} \sum_{i=1}^{N} Dist_{rt}({\textit{box}}_i,{\textit{R}}_i,{\textit{T}}_{i}, {\textit{P}}_{i})
\label{equ:main_objective_old}
\end{aligned}
\end{equation}
where the objective is constrained by the condition that the sequential bounding boxes $\textit{BOX}$ have a smooth trajectory, and the heading of each $\textit{box}_i$ is consistent with trajectory as well. Furthermore, given the initial box state $\textit{box}_i^0$, the final state of $\textit{box}_i$ can be estimated with optimized ${\textit{R}}_i$ and ${\textit{T}}_i$. Instead of finding ${\textit{R}}_i$ and $\textit{T}_i$ as the intermediate state, we can optimize each variable of $\textit{box}_i$ (i.e. $x_{i}$, $y_{i}$, $z_{i}$, $\alpha_{i}$, $\beta_{i}$, $\gamma_{i}$) directly for simplicity. Therefore, we can get our final objective function, as Equ. \ref{equ:main_objective}. The function $\textit{Dist}$ and each constraint will be thoroughly explained in the upcoming section. 

\begin{equation}
\begin{aligned}
\underset{\textit{box}_i}{\arg\min} \frac{1}{N} \sum_{i=1}^{N} Dist({\textit{box}}_i, {\textit{P}}_{i})
\label{equ:main_objective}
\end{aligned}
\end{equation}

%% file: registration.tex
In order to optimize the variables ($x_{i}$, $y_{i}$, $z_{i}$, $\alpha_{i}$, $\beta_{i}$, $\gamma_{i}$), we need to properly encode them in a continuous function so that they are differentiable. The modeling of the objective function and each constraint and its optimization will be explained in detail.

\subsection{Modelling}\label{sec:xxxxxx}
The main objective function (Equ. \ref{equ:main_objective}) aims to align bounding box $\textit{box}_i$ with its corresponding points as closely as possible while ensuring that these points are enclosed within $\textit{box}_i$. Hence, it can be decomposed into two terms: \textit{closeness} $\mathbbmsl{L}_c$ and \textit{enclosure} $\mathbbmsl{L}_e$. The \textit{closeness} $\mathbbmsl{L}_c$ measures the distance between $\textit{BOX}$ and $\textit{SEQ}$, while the \textit{enclosure} $\mathbbmsl{L}_e$ ensures that all points lie inside their corresponding boxes. However, the $\mathbbmsl{L}_c$ term may not always be zero due to the constraints imposed by the smooth trajectory on the pose (location and orientation) of $\textit{box}_i$, and like when the objects are occluded and their point clouds are located inside the bounding box without touching any of its surfaces.

The objective function is constrained by the condition that all $\textit{box}_i$ have a smooth trajectory, and the heading of each $\textit{box}_i$ is consistent with the trajectory as well. The constraints can be depicted with two additional terms: $\textit{smoothness}$ $\mathbbmsl{L}_s$ and $\textit{alignment}$ $\mathbbmsl{L}_a$. The $\textit{smoothness}$ $\mathbbmsl{L}_s$ is used to represent how smooth the trajectory is, while $\textit{alignment}$ $\mathbbmsl{L}_a$ is used to evaluate how well each $\textit{box}_i$ aligns with the trajectory. These four terms will be elaborated on comprehensively in the rest of the section. \\

\subsubsection{\textit{Closeness} $\mathbbmsl{L}_c$}

The main idea of the $\textit{Closeness} \mathbbmsl{L}_c$ term is to keep the $\textit{box}_i$ in close proximity to its corresponding point cloud. To achieve this, we assume that the front/back face of the bounding box is perpendicular along the x-axis, the left/right side is perpendicular along the y-axis, and the top/bottom side is perpendicular along the z-axis. The distance between point $p_i^j$ and the front/back face of $\textit{box}_i$ can be denoted as $(D_{x}(p_j^i, box_{i})$ \footnote{$(D_{x}(p_j^i, box_{i})$ refers to the distance between point and plane and can be easily calculated with equation $(D_{x}(p_j^i, box_{i})=\frac{|a*x_i + b*y_i + c*z_i +d|}{\sqrt{a^2+b^2+c^2}}$, in which $a,b,c,d$ is the parameters for the plane function of one face of $\textit{box}_i$. Given the $box_{i}$, the plane function of its six faces can be easily calculated with rigid geometry transformation. The whole calculation chain is differentiable.}, similarly, we can calculate the $D_y(.)$ and $D_z(.)$. Observed from a certain point of view, the object has at most one visible face along each axis, and there are usually three visible faces. The point clouds should stay near these visible faces. The visible side can be decided by comparing the mass center of the point cloud and the geometrical center of the box, which is simple but feasible given the assumption that the initial bounding boxes are coarsely close to their corresponding point cloud. The top $K$-th closest points are selected to estimate the \textit{Closeness} $\mathbbmsl{L}_c$ instead of using all points, as only boundary points matter. In the end, the $\mathbbmsl{L}_c$ function can be defined as the following Equ. \ref{equ:closeness}.

\begin{equation}
\begin{aligned}
\textstyle \mathbbmsl{L}_c  &\textstyle = \frac{1}{N*K} \underset{i=0}{\overset{N}{\sum}}\underset{j,k,n=0}{\overset{K}{\sum}}(D_{x}(p_{j}^{i}, box_{i})^{2} + \\ &\textstyle D_{y}(p_{k}^{i},box_{i})^{2} + D_{z}(p_{n}^{i}, box_{i})^{2})
\label{equ:closeness}
\end{aligned}
\end{equation}

\begin{figure}[!h]
    \centering
    \includegraphics[scale=0.25]{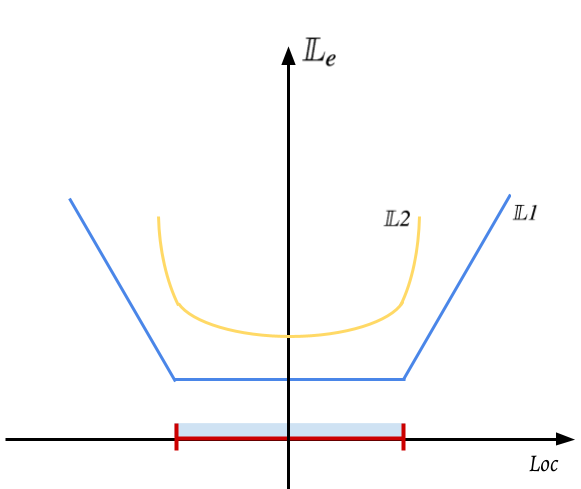}
    \caption{The sketch diagram about the landscape of \textit{Enclosure} $\mathbbmsl{L}_e$ with $\textit{L1}$ and $\textit{L2}$-norm.}
    \label{fig:encloure}
\end{figure}

\begin{figure*}[!h]
\centering
	\subfigure[Visualization]{
	\includegraphics[width=0.48\textwidth]{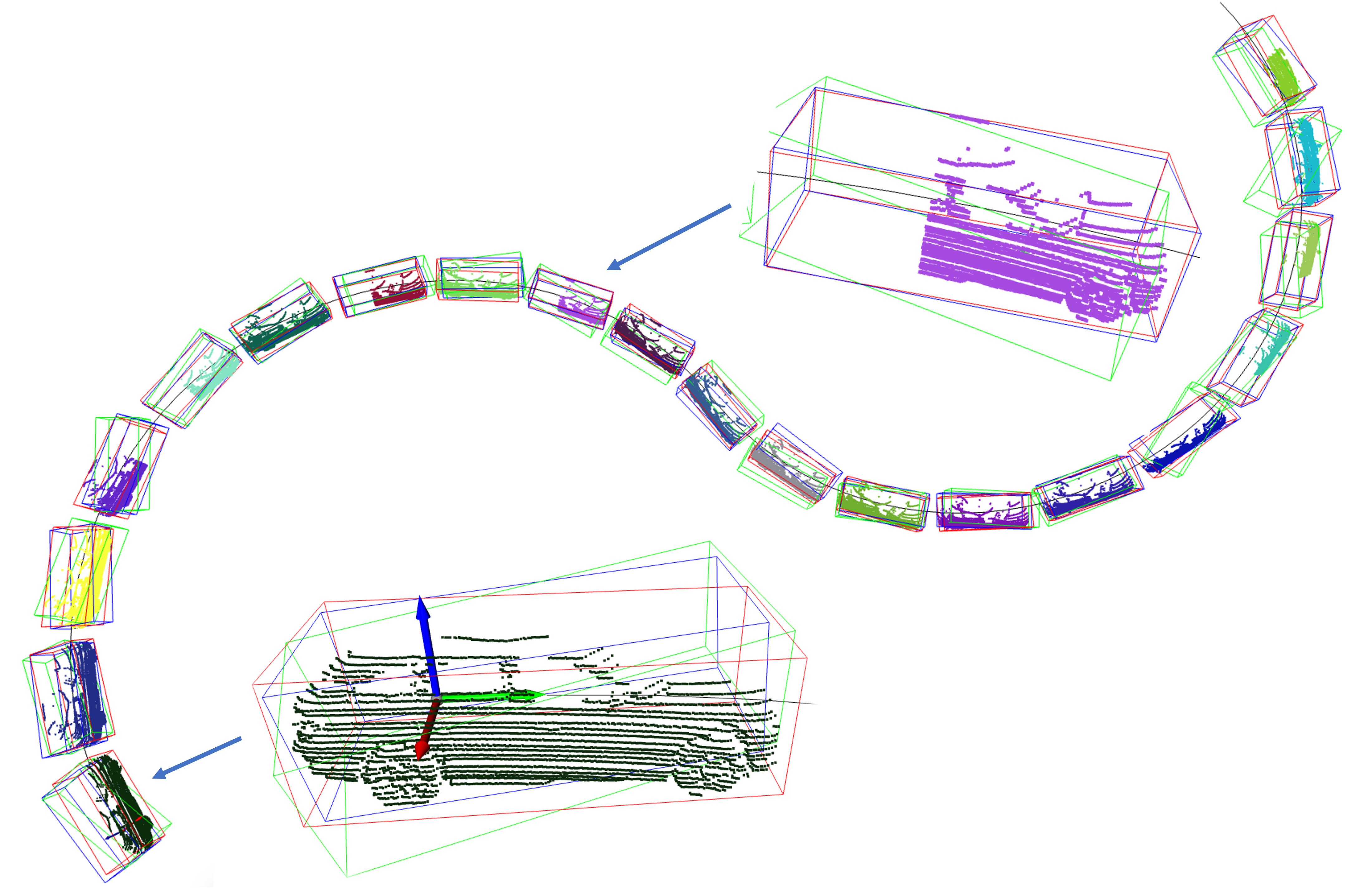}}
        \hspace{-0.2cm}
        \subfigure[Error comparison]{
        \includegraphics[width=0.48\textwidth]{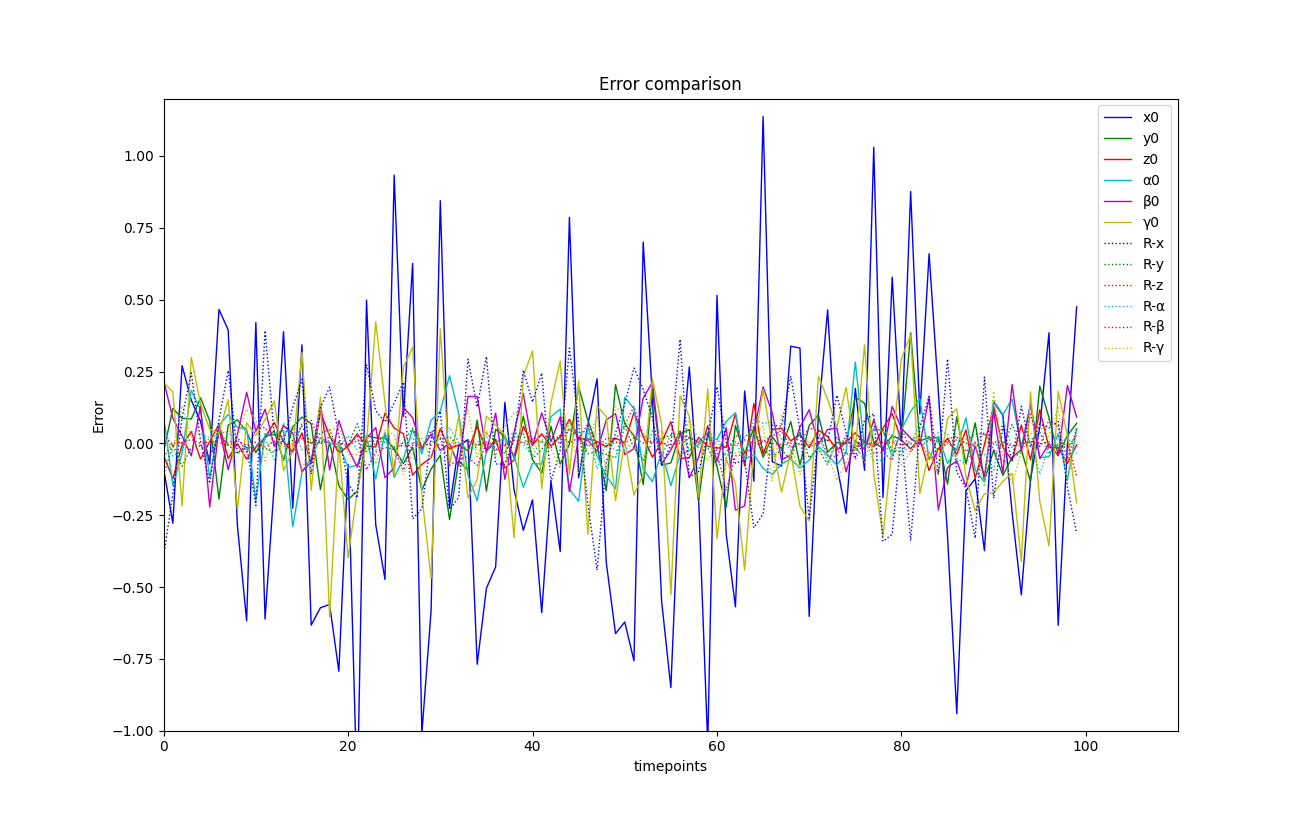}
        }
\caption{The comparison registration results. In Fig. (a), the ground truth bounding boxes are depicted in Red color, the initial bounding box is green, and the registration bounding box is blue color.  Fig. (b) displays the error value. All point clouds and bounding box are temporarily down-sampled by 5 to make plotting clean.}
\label{fig:3d_obj}
\end{figure*}

\subsubsection{\textit{Enclosure} $\mathbbmsl{L}_e$}
A $\textit{box}_i$ should include $M_{i}$ points inside, i.e. $ P_{i} = \{ p_1^i, p_2^i, ....,p_{M_i}^i| i \in N \}$, but it actually only encloses $L_{i}$ points. Therefore, the enclosure $\mathbbmsl{L}_e$ can be defined as the ratio between $L_{i}$ and $M_{i}$, i.e. $L_{i}/M_{i}$. However, this definition is not differentiable with respect to the optimized values ($x_{i}$, $y_{i}$, $z_{i}$, $\alpha_{i}$, $\beta_{i}$, $\gamma_{i}$). To make the enclosure term gradient-traceable to the target values, we represent it as $\textit{L1}$-norm distance between points and six faces of the bounding box, as shown in Equ. \ref{equ:enclosure}. The reason why we choose $\textit{L1}$-norm rather than $\textit{L2}$-norm can be well-explained with their loss landscape, as depicted in Fig. \ref{fig:encloure}. With the $\textit{L1}$-norm, the $\mathbbmsl{L}_e$ is constant and optimal, when the points are located inside the $\textit{box}_i$, so the gradient is thereby zero. As the points move out the $\textit{box}_i$, the $\mathbbmsl{L}_e$ start to increase, accompanied by a non-zero gradient. However, the optimal solution for $\mathbbmsl{L}_e$ with $\textit{L2}$-norm is when points are located in the center of the bounding box, which is not desired towards our objective.

\begin{equation}
\begin{aligned}
\textstyle \mathbbmsl{L}_e = \frac{1}{ N*6}\underset{i=0}{\overset{N}{\sum}}{\underset{k=1}{\overset{6}{\sum}}(\frac{1}{M_i}\underset{j=1}{\overset{M_{i}}{\sum}}{ |D_{k}(p_{j}^{i}, box_{i})| }})
\label{equ:enclosure}
\end{aligned}
\end{equation}

\subsubsection{$\textit{Smoothness}$ $\mathbbmsl{L}_s$}
The sequential bounding boxes $BOX$ have a smooth trajectory with an unknown degree of a polynomial. To model the $\textit{smoothness}$ $\mathbbmsl{L}_s$, we assume that the change in the box's location and orientation is similar between neighboring timestamps, which is similar to a constant-velocity motion model. Therefore, the smoothness $\mathbbmsl{L}_s$ can be defined as Equation \ref{equ:smoothness}.

\begin{equation}
\begin{aligned}
\textstyle \mathbbmsl{L}_s = \frac{1}{N-2}\underset{i=2}{\overset{N-1}{\sum}}|| \varDelta{box_i} - \varDelta{box_{i-1}}||_2
\label{equ:smoothness}
\end{aligned}
\end{equation}
where the $\varDelta{box_i} = [|x_{i+1} - x_{i}| , |y_{i+1} - y_{i}|, |z_{i+1} - z_{i}|, |\alpha_{i+1} - \alpha_{i}|, |\beta_{i+1} - \beta_{i}|, |\gamma_{i+1} - \gamma_{i}|]^T$ describe the pose change between neighbouring timestamps. \\

\subsubsection{$\textit{Alignment}$ $\mathbbmsl{L}_a$}
The heading of the object or $box_i$ is along its x-axis (forward/backward) and it should align with the direction of the trajectory. The orientation vector unit $O_i$ should be close to the movement vector unit of the object $U_i$. The heading alignment $\mathbbmsl{L}_a$ can be defined as the following Equ. \ref{equ:heading_align}. Before we calculate the $L_{alignment}$, each $\alpha_{i}$ is wrapped into $[-\pi/2, \pi/2]$
\begin{equation}
\begin{aligned}
\textstyle \mathbbmsl{L}_a = \frac{1}{N-1}\sum_{n=1}^{N-1}||O_i - U_i||_2
\label{equ:heading_align}
\end{aligned}
\end{equation}
where the $O_i$ = $[\cos(\beta_i)\cos(\gamma_i), \cos(\beta_i)\sin(\gamma_i), -\sin(\beta_i)]^T$ \footnote{The $O_i$ is calculated with the Euler angles-based rotation matrix, for example, a roll about the x-axis is defined as $ R_{x}(\alpha_i)= \tiny \begin{bmatrix} 1&0&0\\ 0&\cos{\alpha_i}&-\sin{\alpha_i}\\0&\sin{\alpha_i}&\cos{\alpha_i} \end{bmatrix}$, simliarly, we can calculate the $R_{y}(\beta_i)$ and $R_{z}(\gamma_i)$, and $O_i=R_{z}(\gamma_i)R_{y}(\beta_i)R_{x}(\alpha_i)[1,0,0]^T$}, and $U_i$ = [$(x_{i+1} - x_{i})/diag$, $(y_{i+1} - y_{i})/diag$, $(z_{i+1} - z_{i})/diag]^T$, ($diag=\sqrt{(x_{i+1} - x_{i})^2+(y_{i+1} - y_{i})^2+(z_{i+1} - z_{i})^2}$).

\subsubsection{Total loss function}
We form the total loss function as a sum of all the above loss terms with an augmented Lagrange multiplier, seen in Equ. \ref{equ:total_loss}. Here, the $\delta$, $\omega$, $\epsilon$, and $\theta$ are the weights and multiplier for each term.

\begin{equation}
\begin{aligned}
\textstyle \mathbbmsl{L}_T = \delta*\mathbbmsl{L}_c + \omega*\mathbbmsl{L}_e +  \epsilon*\mathbbmsl{L}_s +  \theta*\mathbbmsl{L}_a
\label{equ:total_loss}
\end{aligned}
\end{equation}

\subsection{Optimization}

Newton's optimization method \cite{NoceWrig06} can be used to find our solution, as our final loss function (Equ. \ref{equ:total_loss}) is differentiable to each optimized variable, and the gradient descent is also proved to be more efficient than classical ICP \cite{ICP_sgd}. The optimization iteration step is shown as Equ. \ref{equ:newtow_method}. The starting values for iteration are the initial pose of 3D bounding boxes. 

\begin{equation}
\begin{aligned}
\begin{cases}
   \textstyle x^{k+1}_{i} = x^{k}_{i}  - 	\frac{\mathbbmsl{L}_T'(x^{k}_{i})}{\mathbbmsl{L}_T''(x^{k}_{i})} ,     y^{k+1}_{i} = y^{k}_{i}  - 	\frac{\mathbbmsl{L}_T'(y^{k}_{i})}{\mathbbmsl{L}_T''(y^{k}_{i})} 
    \\
   \textstyle z^{k+1}_{i} = z^{k}_{i}  - 	\frac{\mathbbmsl{L}_T'(z^{k}_{i})}{\mathbbmsl{L}_T''(z^{k}_{i})} ,     \alpha^{k+1}_{i} = \alpha^{k}_{i}  - 	\frac{\mathbbmsl{L}_T'(\alpha^{k}_{i})}{\mathbbmsl{L}_T''(\alpha^{k}_{i})}
    \\
  \textstyle  \beta^{k+1}_{i} = \beta^{k}_{i}  - 	\frac{\mathbbmsl{L}_T'(\beta^{k}_{i})}{\mathbbmsl{L}_T''(\beta^{k}_{i})} ,    \gamma^{k+1}_{i} = \gamma^{k}_{i}  - 	\frac{\mathbbmsl{L}_T'(\gamma^{k}_{i})}{\mathbbmsl{L}_T''(\gamma^{k}_{i})} 
    
  \end{cases}
\label{equ:newtow_method}
\end{aligned}
\end{equation}

\subsection{Speeding-up}

The computation of $\mathbbmsl{L}_T$ involves going through each point in the point cloud streams, and given that there could be a large number of points in one bounding box, the optimization process can be quite slow. To speed up the computation, we first down-sample the point clouds using farthest point sampling \cite{qi2017pointnetplusplus}, which can significantly reduce the number of points while preserving the structure of the object. Newton's iteration with the second-order gradient can find the optimal iteration step in Equ. \ref{equ:newtow_method}, but the second-order gradient requires a lot of extra computation. Therefore, we use quasi-Newton methods, specifically, the LM-BFGS algorithm \cite{limited_lbfgs}, for computing and updating the gradients to save computation time.

%% file: experiments.tex
\subsection{Setup details}

\begin{figure*}[h]
\centering
	\subfigure[Visualization]{
	\includegraphics[width=0.4\textwidth]{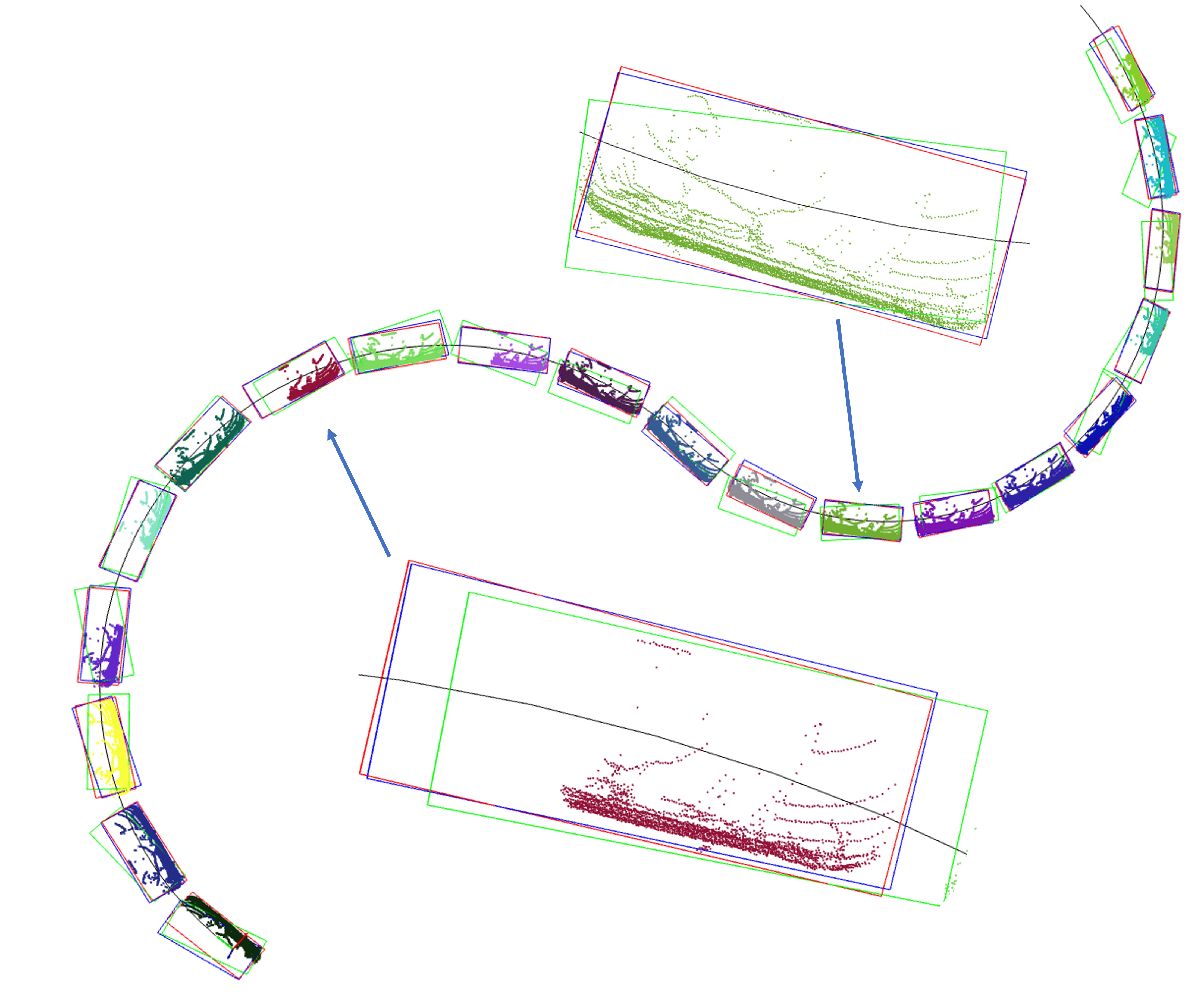}}
        \hspace{-0.1cm}
        \subfigure[Error comparison]{
        \includegraphics[width=0.55\textwidth]{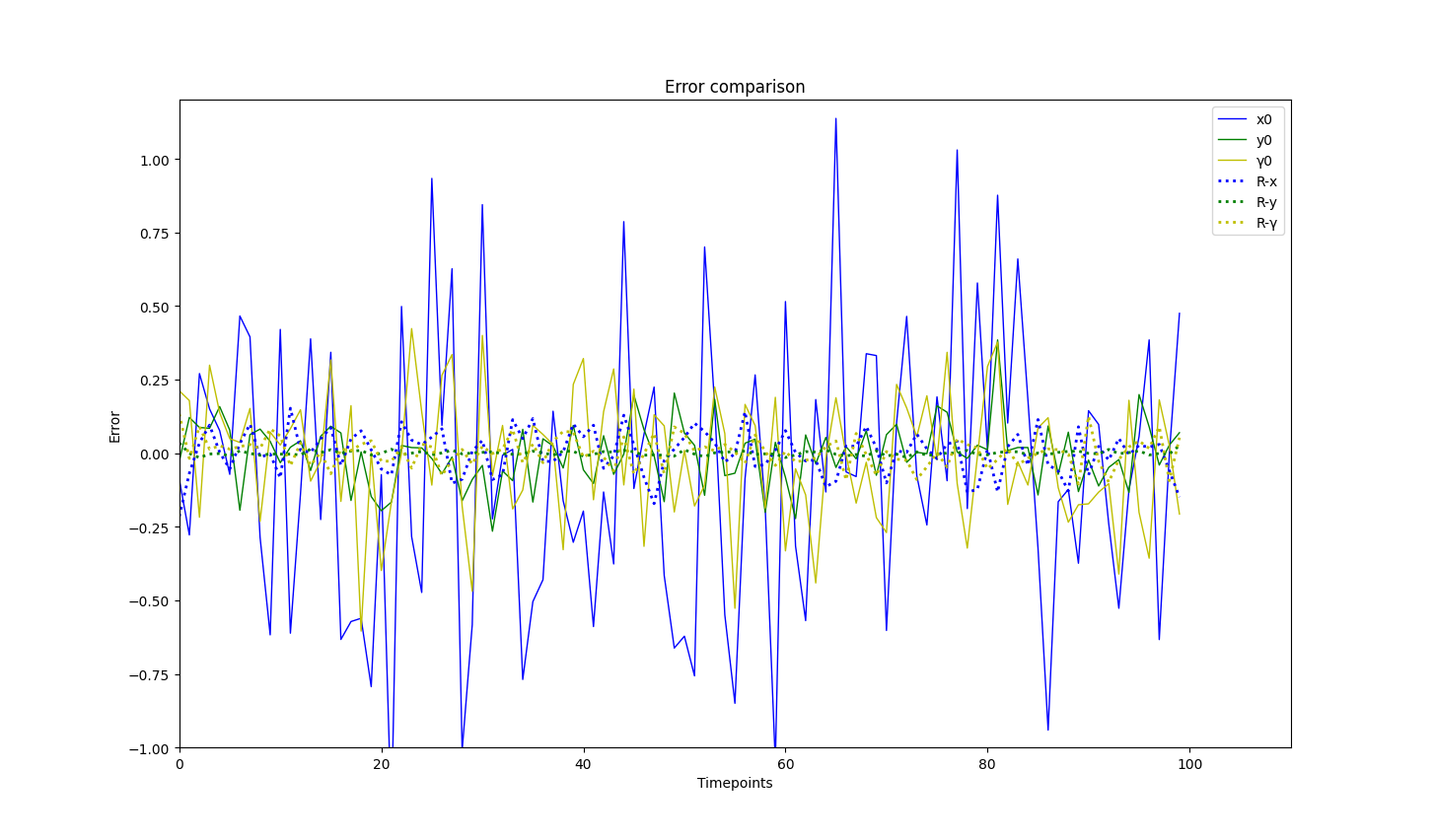}
        }
\caption{The comparison registration results. In Fig. (a), the ground truth bounding boxes are depicted in Red color, the initial bounding box is green, and the registration bounding box is the blue color.  Fig. (b) displays the error value.}
\label{fig:2d_obj}
\end{figure*}

\begin{figure}[]
\centering
	\subfigure[Left-right view.]{
	\includegraphics[width=0.22\textwidth]{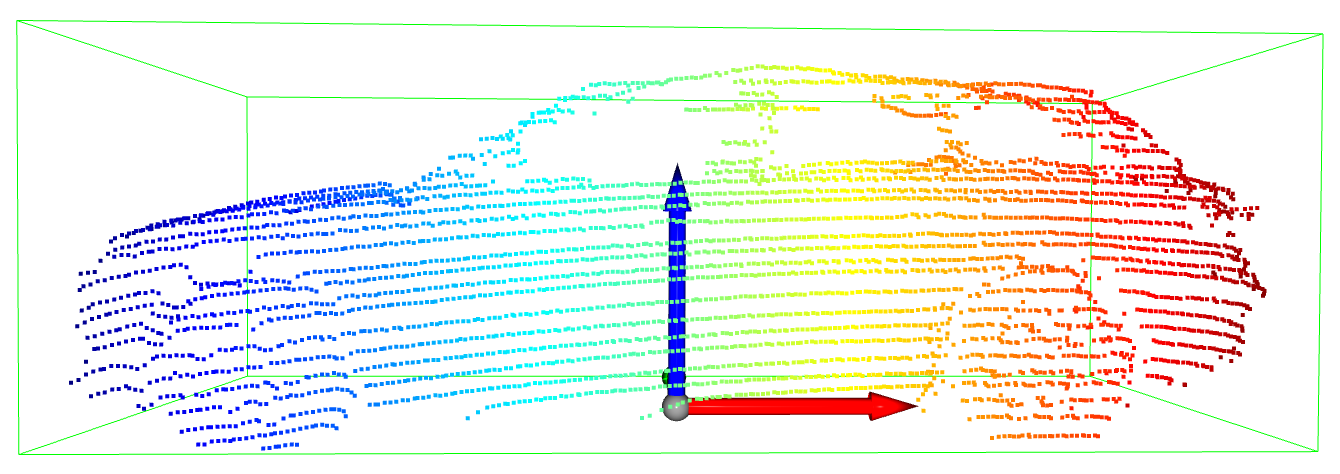}}
	\hspace{0.1cm}
	\subfigure[Top-down view]{
	\includegraphics[width=0.18\textwidth]{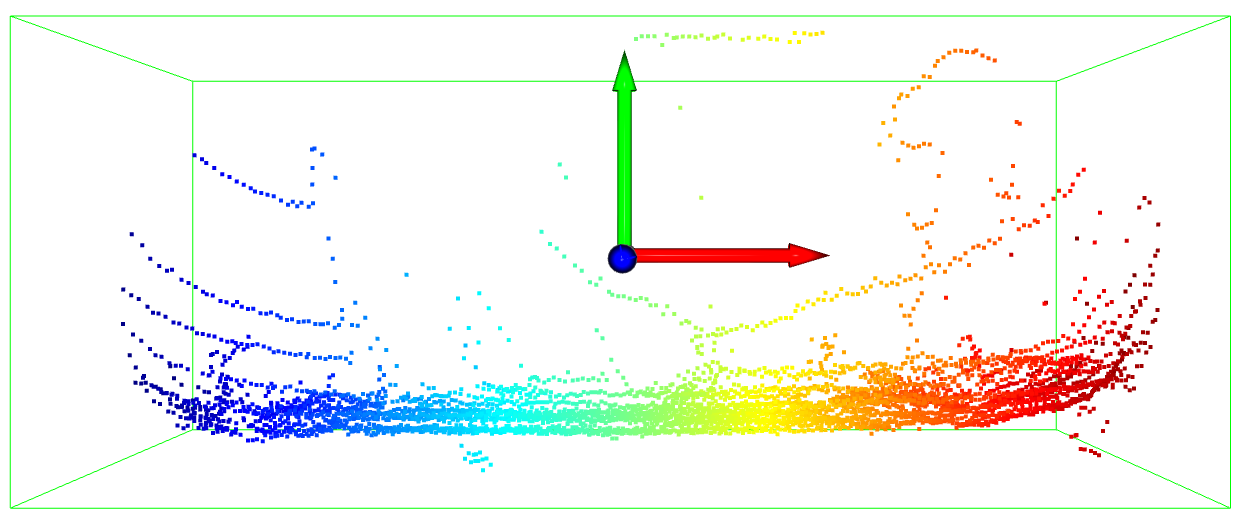}}
\caption{A visualized point cloud example of an SUV for the experiment. We give the (a) left-right and (b) top-down views of the example.}
\label{fig:demo_pc}
\end{figure}

We validate our approach with a simulated point cloud, 3D bounding box, and controlled trajectory. The point cloud and bounding box can be seen in Fig. \ref{fig:demo_pc}, which is an SUV scanned by a LiDAR from a side view. The experiment is conducted in two experimental settings. One is for sequential 3D bounding boxes registration, and we assume that the vehicle is constantly moving forward, and keep turning (yawing) in a half cycle around with rolling and pitching simultaneously, like a vehicle driving on a mountainous and uneven road. The trajectory function is as Equ. \ref{equ:trajectory}. The other setting is for sequential 2D bounding boxes registration and it is a simplified scenario with the assumption that the vehicle is moving on a flat area and detected from bird-eye view (BEV), and $\alpha(t)=0$ (no rolling) $\beta(t)=0$ (no pitching). This is a common assumption in the context of autonomous driving \cite{Vision-Centric-BEV}. For both experiments, we randomly block some areas of the SUV point cloud during the vehicle's movement, as if it were occluded by other vehicles in the real world.

\begin{equation}
\begin{aligned}
\begin{cases}
   \textstyle x(t) = x(t-1) + v*\varDelta{t}
   \\
   \textstyle \alpha(t) = \alpha(t-1) +  sign(\cos(\frac{t*2\pi}{T}))*\frac{\pi}{6*T}
   \\
   \textstyle \beta(t) = \beta(t-1) +  sign(\cos(\frac{t*2\pi}{T}))*\frac{\pi}{3*T}
   \\
   \textstyle \gamma(t) = \gamma(t-1) + sign(\sin(\frac{t*2\pi}{T}))*\frac{2*\pi}{T}
  \end{cases}
\label{equ:trajectory}
\end{aligned}
\end{equation}

\subsection{3D bounding boxes}
The trajectory of 3D object's movement can be found in Fig. \ref{fig:3d_obj}(a), and each state within the trajectory includes the six degrees of freedom. The error comparison before and after registering sequential bounding boxes with the point cloud streams can be found in Fig. \ref{fig:3d_obj}(b). The experimental results show that our optimized bounding boxes in green color are much closer to the ground truth (red), given poor initial boxes in green color, especially for these bounding boxes in the middle. However, for the box at the very beginning and the last end, our approach fails (seeing the zoomed bounding box at the bottom of Fig. \ref{fig:3d_obj}(a)). This is because there is no historical information for the first time point, and no future information for the last time point, which means that $\textit{Smoothness}$ $\mathbbmsl{L}_s$ (Equ. \ref{equ:smoothness}) and $\textit{Alignment}$ $\mathbbmsl{L}_a$ (Equ. \ref{equ:heading_align}) are not available for generating gradients, and there are no sufficient constraints enforced on optimizing them. From Fig. \ref{fig:3d_obj}(b), we can observe that the dotted lines are more concentrated around zero than solid lines, which indicates that our approach can significantly reduce the error between initial bounding boxes and ground truth. The error difference can also be found in Table \ref{tab:2d_3d_comparison}, from which we infer that our approach can effectively optimize each variable and reduce error to 30\% and improve the average IoU from 0.573 to 0.816. Each 3D bounding box has seven parameters for optimization, and as the trajectory increases, the dimensions increase as well, making the optimization more challenging, especially for the roll $\alpha_i$, which has only one loss term imposed on it. Sliding window techniques can be adapted to alleviate high-dimensional issues, and fixed-length windows with a low number of optimized parameters allow for stable optimization.

\begin{table}[h]
\resizebox{0.48\textwidth}{!}{
\begin{tabular}{c|c|c|c|c|c|c|c} \hline
     Item & x & y & z & $\alpha$ & $\beta$ & $\gamma$ & IoU  \\ \hline
     Initial  & 0.392 & 0.124 & 0.082 & 0.083 & 0.1 & 0.18  &  0.573/0.618  \\ \hline 
       3D  &  0.121 & 0.049 & 0.012 & 0.031 & 0.034 & 0.053 & 0.816   \\ \hline
       2D  &  0.054 & 0.006 & n/a & n/a & n/a & 0.032 & 0.896  \\ \hline
\end{tabular}}
\caption{Performance comparison between 2D/3D bounding boxes. The second row is about the mean error between initial bounding boxes and ground truth boxes, and the two IoU values (0.57/0.62) representing 2D/3D IoU respectively (they are put together to keep the table clean). The third rows show the mean error between 3D bounding boxes and ground truth, the last row is for 2D bounding boxes. Unit for $x,y,z$ is meter, and unit for $\alpha, \beta, \gamma$ is the radian.}
\label{tab:2d_3d_comparison}
\end{table}

\subsection{2D bounding boxes in BEV}
2D object detection from BEV is also an important topic and has been widely used for autonomous driving \cite{Vision-Centric-BEV}. We applied our approach to this simplified scenario as well. The problem becomes the registration between 2d bounding boxes (no rolling, pitch, and height) and 2D point cloud (no z value), and our optimized variables change from 6 ($x_{i}$, $y_{i}$, $z_{i}$, $\alpha_{i}$, $\beta_{i}$, and $\gamma_{i}$) to 3 ($x_{i}$, $y_{i}$, $\gamma_{i}$). We used a similar trajectory as in the previous part, shown in Fig. \ref{fig:2d_obj} (a). We can visually tell that the blue bounding boxes are very close to the red ones and our approach can achieve better performance in the 2D tasks than the 3D task. From the table \ref{tab:2d_3d_comparison}, we can see that mean errors for three variables ($x_{i}$, $y_{i}$, $\gamma_{i}$) in 2D bounding boxes are much smaller than the previous task, and the proposed approach can increase the IoU from 0.618 to 0.896 (40\% improvement). The reason why our approach performs better is mainly due to the low dimensionality of variables. This task has only half of the total parameters in previous tasks, and each variable can be optimized by multiple loss terms with strong constraints.